# Multi-Label Clinical Text Eligibility Classification and Summarization System

Surya Tejaswi Yerramsetty    Almas Fathima

*Abstract*— Clinical trials are at the heart of all medical advances as they help us better understand and improve human health and the healthcare system. They play a fundamental role in determining new ways to detect, prevent, or treat a disease, and it's essential that clinical trials include people with a variety but appropriate medical histories and backgrounds. In this paper, we propose a system that leverages the power of NLP and LLM models to automate Multi-Label Clinical Text Eligibility Classification and Summarization. This system utilizes a combination of feature extraction techniques like word embedding (word2vec) and entity recognition to identify relevant medical concepts and traditional approaches like count vectorizers and TF-IDF (Term Frequency-Inverse Document Frequency). This paper also explores methods like weighted TF-IDF word embeddings, which combine the strengths of both count-based and embedding-based methods to capture word importance more effectively. Multi-label classification with random forest and SVM models are employed to categorize documents based on eligibility criteria. We explore various summarization techniques (TextRank, Luhn summarization, GPT-3) to summarize the eligibility requirements concisely. Evaluation using ROUGE scores demonstrates the effectiveness of the summarization methods. This system can potentially automate clinical trial eligibility assessment using data-driven methods, thus improving research efficiency.

## I. INTRODUCTION

Clinical trials are fundamental in evaluating new medical interventions and are the cornerstone of clinical research. They are widely used across the world in the healthcare system. They are essential for paving the way towards medical advancement by enabling the development and wide-range testing of novel treatments or new ways to use existing treatments, eventually leading to improved patient care. Clinical trials provide a scientific basis for advising and treating patients even when researchers do not obtain the outcomes they predicted; trial results can help point scientists in the right direction. A critical yet time-consuming hurdle in this process is identifying participants who satisfy and meet the specific eligibility criteria outlined in each trial's protocol. This is a crucial step because the trials are performed on a much smaller sample size than the larger audience that will eventually consume or adopt the new practices. Traditionally, this assessment relies on a manual review of lengthy and intricate clinical text documents of diverse medical histories, a method prone to errors and inefficiencies. "Additionally, if a researcher does not have time to scour through clinical narratives, they may recruit patients who seek out the clinical trials independently or are encouraged to enroll in the study by their primary care physician. These practices can result in selection bias toward certain populations, such as those who are more likely to have a primary care physician or people who have the knowledge and time to search for relevant clinical trials on their own." [1] Therefore, automatic clinical text analysis leveraging NLP and LLM capabilities offers a compelling solution to expedite and enhance eligibility determination and becomes imperative. Automating this process can significantly reduce selection bias, leading to more accurate and unbiased research outcomes.

Natural Language Processing (NLP) and Large Language Models (LLM) provide machines with the ability to understand, interpret, and extract important information from vast amounts of unstructured text data with the capability of simulating human intelligence. By employing advanced algorithms and linguistic patterns, NLP systems can identify key medical concepts, extract relevant data points, and precisely categorize information. With their understanding capability, LLMs, such as GPT-3, further enhance this by providing deep learning models with a nuanced understanding of language and context. These NLP and LLM models can understand and extract complex medical terminology, identify patterns, detect subtle nuances in patient narratives, and generate accurate summaries of large clinical documents. This paper introduces a system that leverages Natural Language Processing (NLP) and Large Language Models (LLM) to streamline the recruitment process for clinical trials by facilitating efficient classification of criteria and comprehension of eligibility criteria from textual data.

The textual data set used in this project is obtained from Track 1 of the 2018 National NLP Clinical Challenges (n2c2), which explored using NLP to identify eligible patients from narrative medical records; this research addresses the limitations of manual review and the potential for bias in participant selection. This paper details the development and evaluation of our data-driven NLP-based system for multi-label clinical text eligibility classification and summarization. We present the utilized dataset, related work methodology, and results, demonstrating the effectiveness of our approach in facilitating efficient and unbiased clinical trial recruitment.

## II. LITERATURE REVIEW

The growing volume of electronic health records (EHRs), clinical notes, and medical data in healthcare has resulted in thorough research in Natural Language Processing (NLP)

.

techniques focusing on developing methods to accurately classify and summarize long medical records, facilitating faster retrieval and analysis of critical information. However, data generation is a huge challenge in the medical field because of its privacy concerns, and data collection is an ongoing challenge. Our project uses an annotated dataset to perform supervised multi-label classification. This challenge dataset was released in 2018, National NLP Clinical Challenges (n2c2), which aimed to answer the question, "**Can NLP systems use narrative medical records to identify which patients meet selection criteria for clinical trials?**" [1]. In this section, we will discuss various works done to answer the above question.

In their paper, the authors of the dataset explained the manual annotation procedure followed for all 288 records; each patient's record indicates whether the patient satisfies a set of selection criteria. The standards were gathered from actual research published on ClinicalTrials.gov. Most patients in the dataset have heart disease risk factors, and all of them have diabetes. "Each annotator examined every patient according to all thirteen criteria and categorized each as met, not met."[1]. However, based on the available data, our project classified the patients according to four major criteria: ABDOMINAL, ADVANCED-CAD, MAJOR-DIABETES, and CREATININE.

- ABDOMINAL: If the patient has any history of intra-abdominal surgery, small or large intestine resection, or small bowel obstruction.
- ADVANCED-CAD: Taking 2 or more medications to treat CAD, History of myocardial infarction, Ischemia, past or present.
- MAJOR-DIABETES: If the patient is suffering from uncontrolled diabetes.
- CREATININE:  Serum creatinine > upper limit of normal

Many participants in the challenge have worked on rule-based approaches to classify the patients' clinical notes on this dataset. For example, in their rule-based approach, Karystianis and their team [2] hand-crafted 12 dictionaries and 280 rules to classify further and achieved a micro F1 Score of 0.89. Despite the good results, this approach is time-consuming and may not work for all the clinical notes. Moreover, implementing a similar model for different criteria would require a re-work of all the rules crafted. Secondly, Ying Xiong [3] implemented LSTM and CNN models for classification tasks; this system has shown good results for the majority classes with a micro F1-score of 0.85 and poorly performed for the minority classes due to the unbalanced nature of the data. Recently, in 2024, leveraging the advancements in LLM, Michael Wornow [4] proposed a zero-shot classification using the GPT-4 model, showcasing the power of LLMs in text classification. His paper also discussed how to improve the cost efficiency of LLM models.

### III. METHODOLOGY

Our Project is the fusion of classification and summarization tasks. For the researcher using our UI, who is trying to know if the patient fits the criteria or not, along with the criteria, he also gets a justifying summary of why he is eligible, highlighting the key points from the clinical notes, which improves making faster and the informed decisions. In our study, we did a thorough analysis to understand the nature of the data. We tried various combinations of feature engineering techniques like word embeddings, n-grams, TF-IDF and count vectorizers, presence or absence of named entities to classify the problem. To handle the multi-label classification problem, we have used Binary relevance, and Classifier chains approaches on the Random Forest and SVM classifiers. We have leveraged the pre-trained large language model's capability of classifying text with minimal effort by implementing a multi-label zero-shot GPT classifier. For the extractive summarization, we used the Luhn summarizer and extracted keywords from the documents using the TF IDF transformer and gave sentence scores accordingly for the summarization; this implementation has given the best result.  We have also generated an abstractive summary by performing basic prompt engineering by incorporating chained prompts to get a precise and concise summary relevant to the classification result using the Open AI GPT -3 model. The overall architecture is shown below in Figure 1. This section will be subdivided into Data, classification, summarization, and User Interface, further discussing a detailed description of all the methods used.

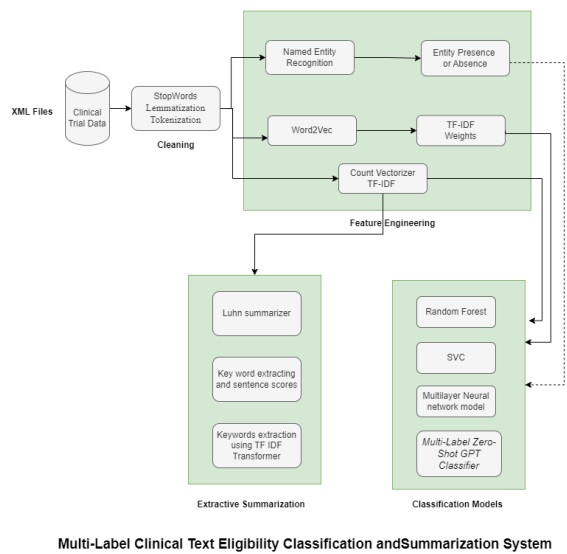

Figure1: Core architecture

### A. Data

The dataset comprises 288 patient records, each annotated at the patient level as "met" or "not met" for all the criteria. We have used 280 records throughout the model implementation for training and testing purposes. The remaining 8 records are used to validate our UI and classification models' performance for the unseen data.

**Data Preprocessing:**

Since we are dealing with textual data in our project, it is important to preprocess it. Next, the data was cleaned and converted into a structured format to analyze the patterns. The individual patient records were cleaned using a series of packages running on top of Python, such as the Natural Language Toolkit (NLTK) that provides stop-word removal, lemmatizing, tokenization, and other data cleanings like lowercase transformation and punctuation removal.

Regular expressions have been exhaustively used to remove unwanted characters like date, time (as it is de-identified notes, this data is meaningless), and a few special characters. This helps machine learning algorithms run smoothly and quickly by reducing the number of input features.

- Stop-word elimination: Removal of the most common words in a language that is not helpful and generally adds a lot of disturbance. Text containing many stop words may cause bias in the machine learning models. By removing stop words, more focus will be on the important information.
- Lemmatizing enhances system accuracy by returning a word's base or dictionary form. Unlike stemming, lemmatization doesn't truncate words without understanding their context. Hence, it is used for meaningful word extraction.
- Tokenizing: Divide the text input into tokens, such as phrases, words, or other meaningful elements, resulting in a sequence of tokens.

**Understanding data:**

Understanding the textual data is crucial before implementing any NLP task. For the target label ADVANCED-CAD, we created a corpus with the records of patients meeting this criterion and calculated the most frequent words. For instance, words like "chest," "pain," and "cardiac" seem to be correct as they are the common complaints of a patient suffering from heart disease. After this, we cleaned our data further because of the noise, like numerical data and the word "mg" (the measuring unit of lab value). Additionally, as visualized in Figure 2, the nature of the classification problem is unbalanced because of the imbalance in classes, suggesting accuracy might not be a good choice as a performance metric.

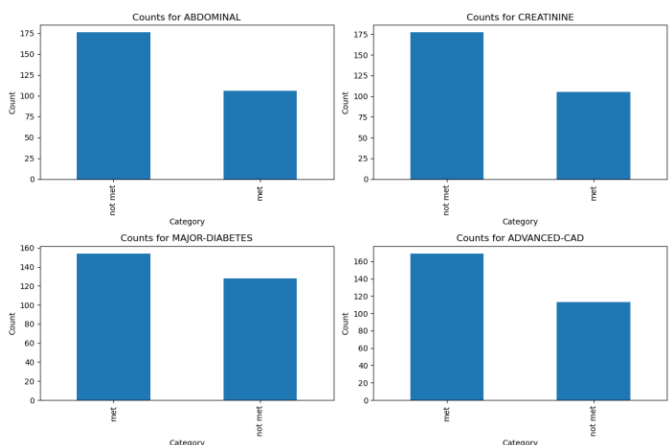

*Figure 2: Distribution of classes in each target variable*

We classified a single target label using a Random Forest model to identify our model's most important features. This approach is known for capturing prominent features in the data. A word cloud visualization (Figure 3) of these prominent features reveals terms like 'Ischemia,' 'myocardial,' and 'coronary.' This aligns precisely with the criteria for manual annotations considered by the authors of the dataset [1], which included medications for **coronary** artery disease (CAD), history of **myocardial** infarction, and presence or absence of **ischemia**.

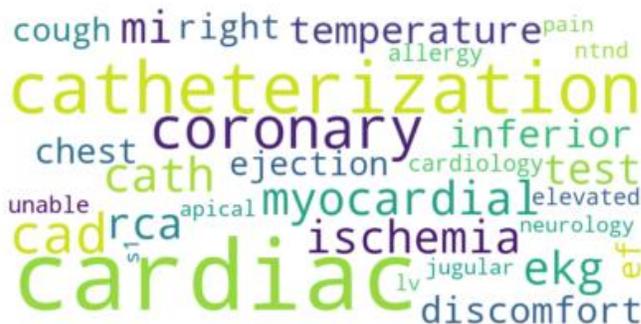

*Figure 3: Random Forest Feature importance Word cloud for ADVANCED-CAD criteria*

### B. Multi-Label Classification

*What makes our problem a multi-label task?*
Unlike multiclass problems, a single data point can simultaneously be associated with multiple labels (one or more classes of target variables) in a multi-label classification. In other words, the predicted classes are **not mutually exclusive**. For our project, the patient might be eligible for different criteria based on his medical history and health issues. We have used different approaches to classify multi-label, such as a Binary Relevance, Multi-Output Classifier, and a Classifier chain.

*Binary Relevance*: In this approach, a multi-label classification problem with L labels is transformed into a separate L single-label binary classification problem, with each classifier predicting the presence or absence of one class for a given data point. The union of the predictions of all the individual classifiers is then taken as the multi-label output.

*Classifier Chains:* As the name suggests, this approach uses a chain of classifiers, with each classifier using the predictions of all the previous classifiers, taking label correlation into account. The total number of classifiers equals the number of labels, like the Binary Relevance approach. For instance, for the data with features (x1, x2 ..... xn) and multiple target labels (y1, y2,y3,y4)., this approach performs a chain of classifications first for y1, then takes the output of the first classification and y2 and performs a classification for this and so on. [4]

*Multi-Output Classifier*: Python's library sklearn has a function to deal with the data having multiple targets. "This strategy consists of fitting one classifier per target. This is a simple

strategy for extending classifiers that do not natively support multi-target classification." [5]

We evaluated the performance of these approaches, and our evaluation metrics (micro average F1 Score) revealed that the Classifier Chain model achieved the best performance. However, the Binary Relevance and Multi-Output Classifiers also produced comparable performance. This might be due to the ability of Classifier Chains to capture label dependencies. Unlike Binary Relevance, which treats labels independently, Classifier Chains allow each classifier to consider the predictions of previous stages, potentially leading to more accurate predictions when labels have inherent relationships.

**Feature Engineering:**

The multi-label nature of data presented a challenge due to class imbalance. We have performed feature engineering to capture textual data's underlying semantics and relationships to improve its performance.

*Named Entity Recognition:* This is crucial in extracting valuable information from the text data. We have implemented a transformer model, 'Clinical-AI-Apollo/Medical-NER', which is pre-trained on a large corpus of medical data. This has successfully identified 18 unique entities in the text corpus, such as Sign_Symptom, Disease_disorder, Diagnostic_procedure, lab_value, etc.
We created informative features with these entities by calculating each entity's presence (0) / absence (1) within the text data point and using them as features in our classification model. This approach is useful as it does not capture redundancy in the data.

*TF-IDF:* It is a statistical measure used to score the importance of a word (term) in any content from a collection of documents based on each word's occurrences; it checks how frequent the keyword is in the corpus and its inverse how rare it is across the document. We have used Python's sci-kit-learn library, which helps convert text into numerical vectors. In the TFIDF vectorizer argument, we have specified the n-gram range as (1,2), which will consider unigrams and bi-grams and calculate the TD IDF weights within the text. This will help to capture the textual patterns not only between the words but also with features.

*Count Vectorizer:* Unlike TF-IDF, this will assign weights based on the word's frequency. We have opted for TF-IDF weights throughout our project as this focuses on prioritizing the occurrence of important words, which proved successful, as shown in Table 1.

*Word Embeddings:* This technique represents words in a vector (Numerical representation of text). We have employed the pretrained Word2Vec model with 100 dimensions to capture the semantic relationships within words in the text. However, to derive more prominent features for our model, we calculated the weighted embeddings by taking their average (which represents the single vector, which represents the overall semantic meaning of the word) and TF-IDF weighted embeddings, which give higher weights to the word that is more frequent in the text (TF) and how rare it is across the text (IDF). We have performed this approach to leverage the advantage of both TF IDF and Word2vec. While one captures the word importance, the other will capture the semantic relationships for better performance. We have tried choosing various combinations as features of our model; this approach has improved our classification model compared to all the techniques, as shown in Table 1 of the results.

**Machine Learning Models:**

This project tackles a supervised learning task with multi-label classification. After reviewing related work on multi-label classification on similar data types, we opted for random forest and support vector classifier models. We have also implemented a neural network perceptron model with sigmoid activation and leveraged the LLM capability to classify using a Multi-Label Zero-Shot GPT Classifier.

Support vector machines are natively equipped to perform binary classification tasks and do not have the inherent ability to perform multi-label or multi-class classification. An SVM model uses kernel functions to learn a decision boundary between two classes. The decision boundary, a hyperplane, can separate data in spaces with more than two dimensions. A multi-label classification problem that involves assigning multiple labels to an instance can be converted into many classification problems with the help of Scikit-learn's MultiOutputClassifier.

The Random Forest classifier is an ensemble model comprised of multiple decision trees for classification tasks. It is a meta-estimator that fits multiple decision tree classifiers built using bagging and random feature selection. Bagging involves creating multiple subsets of the training data for each decision tree, and at each tree growth, a subset of features is randomly selected, thus introducing randomness. Using these techniques, the Random Forest classifier improves performance and reduces the model's variance. After comparing the results, this model performed well, with less overfitting to the training data and explaining more variability to the unseen data.

We have also implemented a basic multi-layer neural network model with four layers and a sigmoid activation in the final layer to perform the multi-label classification. However, this approach assesses the performance of neural networks for this data, and the model's performance could have been improved by using more advanced deep learning models.

*Multi-Label Zero-Shot GPT Classifier:* The latest integration of LLMS in scikit-llm python's library has made it easier to leverage GPT's natural language understanding capabilities. With no feature engineering, we have just vectorized the data using a GPT vectorizer and fed it to the classification model. The performance of this model is like that of traditional machine learning approaches that utilize more prominent features after extensive feature engineering.

**Evaluation Metrics and Results:**
We have chosen micro-average scores as our primary performance indicator based on our understanding of the data and the multi-label classification characteristic with unbalanced classes. In the case of micro-averaging precision and recall, all the individual True Positives, True Negatives, False Positives, and False Negatives for each class are summed up, and their average is taken. The micro-average F1 score is the harmonic of the micro-averaged precision and recall.

However, the best model, i.e., a chain of random forest classifiers leveraging a wide range of features (highlighted) in Table 1 below, is not selected solely by considering the micro averages; we have evaluated the model based on training and testing results to ensure it isn't overfitted, as this hinders its performance on unseen data. Also, we have considered the minority class performance, which is consistent with the distribution of classes in the dataset.

| Features | Model | Precision | Recall | F1 score |
|---|---|---|---|---|
| TF IDF | Multi-Output RF | 0.66 | 0.66 | 0.66 |
|  | Multi-Output SVC | 0.6 | 0.62 | 0.61 |
| NER (present/absent) +TF IDF | RF | 0.64 | 0.75 | 0.69 |
|  | SVC | 0.59 | 0.74 | 0.61 |
|  | Binary Relevance SVC | 0.65 | 0.78 | 0.71 |
| TF IDF weighted embeddings +TF IDF (processed text) | SVM | 0.67 | 0.7 | 0.69 |
|  | RF | 0.68 | 0.71 | 0.69 |
|  | Binary Relevance | 0.66 | 0.65 | 0.66 |
|  | Classifier chains: SVC | 0.69 | 0.72 | 0.72 |
|  | Classifier chains: RF | 0.75 | 0.84 | 0.83 |
|  | Neural Network Perceptron | 0.62 | 0.9 | 0.72 |
| Count Vectorizer + NER | RF | 0.63 | 0.7 | 0.67 |
|  | SVM | 0.63 | 0.8 | 0.7 |
| GPT Vectorizer (Processed_text) | Multi-Label Zero-Shot GPT Classifier | 0.48 | 0.86 | 0.62 |

*Table 1: Micro average Precision, Recall, and F1 Scores of all the classification models.*

### C. Summarization

Summarization can be broadly classified into two types extractive and abstractive. In extractive summarization, the original text is not paraphrased; instead, the same sentences are used based on their scoring to generate summaries. In abstractive summarization, the key information or idea is captured, and the summary is generated accordingly. It involves a deeper understanding of the text relative to its domain. In our project, we generated summaries using both extractive and abstractive summarization models.

**Extractive Summarization:**
Luhn Summarization: Luhn Summarization is a basic summarization technique in which each word is prioritized based on the number of times it occurs in a text, with higher weights assigned to the words present at the beginning of the document. Luhn's score for each sentence is calculated based on these important words. Summarization is done based on the Luhn score for each sentence. This model performed well in precision, as shown in Table 2, indicating that relevant content is present in summaries. However, there is room for improvement in recall and overall F1 score. Due to their complexity, most clinical text summarizations prefer extracting important keywords first and then doing further analysis based on these features.

Text Summarization using TF-IDF: Term Frequency (TF) measures the number of times the word occurred in a document, and IDF measures the uniqueness of a word in a document. Based on these two factors, the importance of a word is calculated, and words with higher scores are considered more important. The words with higher scores are selected as keywords for each document. Using these keywords, sentences are scored based on their relevance or importance. The sentences with high scores are represented as summaries. Like the Luhn, this model showed high precision but relatively low recall across Rouge-1, Rouge-2, and Rouge-L metrics. It suggested that it generated high-quality summaries for the identified n-grams and common sub-sequences. However, there is room for improvement in capturing more relevant content from the reference summaries.

Text Summarization using Count Vectorizer and TF-IDF Transformer: Count vectorizer converts words into numbers, which is easily interpretable by the machine language. It is basically a matrix where each record represents a document and a word as a column. The number of occurrences of each word is counted and stored. Now, TF-IDF is applied to this matrix to identify unique, important features in a document and assign a score to them. Based on these scores, keywords are extracted. The extracted keywords filter out each sentence based on the relevance to these keywords to extract a summary.

| Model | Rouge Scores | Precision | Recall | F1-Score |
|---|---|---|---|---|
| Luhn Summarizer | Rouge-1 | 1.0 | 0.22 | 0.37 |
| | Rouge-2 | 0.99 | 0.19 | 0.32 |
| | Rouge-L | 1.0 | 0.22 | 0.37 |
| TF-IDF | Rouge-1 | 0.97 | 0.21 | 0.35 |
| | Rouge-2 | 0.95 | 0.17 | 0.29 |
| | Rouge-L | 0.97 | 0.21 | 0.35 |
| Count Vectorizer & TF-IDF Transformer | Rouge-1 | 1.0 | 0.26 | 0.42 |
| | Rouge-2 | 0.96 | 0.20 | 0.33 |
| | Rouge-L | 1.0 | 0.26 | 0.42 |

*Table 2: Rouge performance metrics*

**Abstractive Summarization:**
We generated an abstract summary that justifies the classification output, i.e., "why a patient is eligible for a clinical trial?" with the relevant extracted information from the patient notes. For this implementation, we have leveraged the Lang Chain framework to interact with the Open AI LLM, thus implementing a multi-step prompting process (chain of prompts) to generate a summary. This will help the researcher to see the patient's history related to his eligibility.

Prompt Engineering: We tried and refined our summarization model using a trial-and-error approach. To achieve this, we have implemented a "refine" chain mechanism that comprises two prompts.
- Question Prompt: This is used to extract the relevant information from the patient data.
  "You need to summarize the patient's clinical data {text} focusing on the key factors that make this patient a strong candidate for the clinical trial.
  You can use the following points:
  Chief Complaint:
  Patient Information:
  Medical History:
  Why he can be considered for the clinical trial."
- Refine Prompt: This will refine the above-generated summary based on the prompt specified.
  "Your job is to produce a summary based on the Classification output explaining why the patient can be selected for: {', '.join(decoded_list)}."

D. *User Interface*

We leveraged the Flask app to build a user interface as it is an easy-to-learn lightweight backend framework with minimum dependencies. Using basic HTML templates, we created our UI for the user to pass the input text, in our case, clinical text data. The user input is then routed to the backend functions for cleaning and feature extraction. Finally, the data is classified using the imported pre-trained ML model (.pkl file), as shown in Figure 4. The abstractive summary is generated based on the decoded list classifier output, explaining the patient's eligibility, as shown in Figure 5.

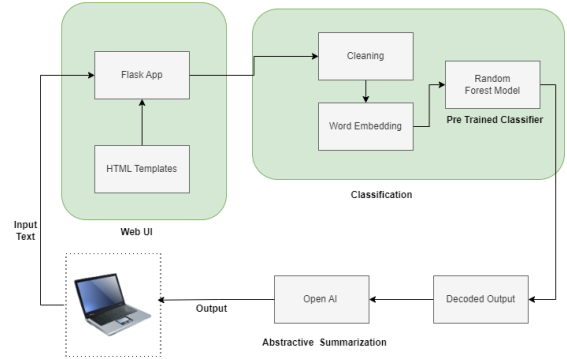

*Figure 4: Architecture for front end*

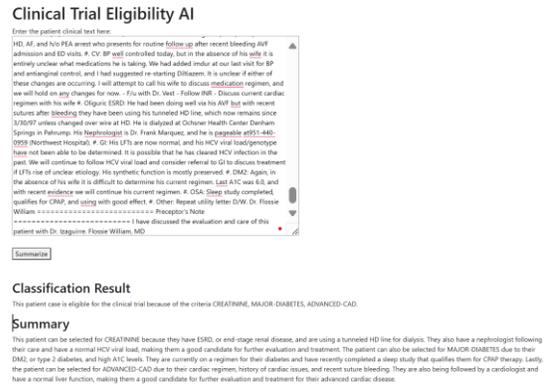

*Figure 5: Final output from UI.*

We implemented an exception-handling mechanism to ensure that a user-friendly message asks the user to pass the actual clinical text when nonclinical data is passed through the UI, as shown in Figure 6.

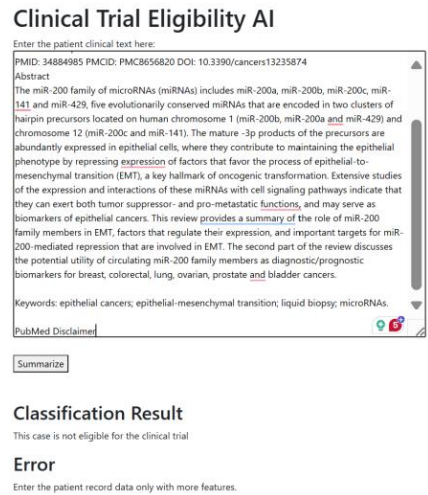

*Figure 6: Result for a medical article.*

## IV. DISCUSSIONS

Our project aim is to develop a clinical screening tool that has been successfully implemented and is working as intended; however, careful consideration could have improved the performance metrics. We have explored ways to improve extractive summarization by incorporating traditional rule-based approaches and custom NER functions to extract relevant and accurate keywords that can summarize efficiently.

For classification, after carefully evaluating the True negatives and false positives cases, it is important to accurately classify for false positives, i.e., when an eligible candidate is wrongly classified as not eligible. To handle this case, we need to design a robust model. This could be achieved by extensive hyperparameter tuning and feature selection as it identifies the optimum parameters at which the model performs effectively.

For abstractive summarization, we have identified an issue with the hallucinating nature of LLM. For instance, the classifier has wrongly predicted that he is eligible for Creatinine criteria while the ground truth is he can be selected for a diabetes trial. The model generated a summary convincingly, stating, "Patient has chronic diabetes this may damage his kidneys in the future so he can be selected for the clinical trial," However, adding the summary feature is valuable as we are not solely relying on model performance. Also, by reviewing the summary, one can easily identify whether the reasoning determines the patient's suitability or not.

## V. CONCLUSION

In conclusion, our project presents a novel data-driven system for automating clinical trial eligibility classification using a multi-label classifier and an Open AI GPT-3 model to generate an abstractive summary. It is a valuable initial screening tool that saves much time for a researcher reviewing lengthy and complicated clinical notes of hundreds of applications. Incorporating this type of tool in pharmaceutical companies will eliminate biases in recruitment processes. This project can be used as a prototype and further enhanced to develop a more robust and sophisticated system. A more user-friendly UI can be created with many other functionalities, even letting the researcher chat with the patient clinical notes. This tool can significantly improve the efficiency and objectivity of clinical trial screening.

## REFERENCES

[1] Stubbs A, Filannino M, Soysal E, Henry S, Uzuner Ö. Cohort selection for clinical trials: n2c2 2018 shared task track 1. J Am Med Inform Assoc. 2019 Nov 1;26(11):1163-1171. doi: 10.1093/jamia/ocz163. PMID: 31562516; PMCID: PMC6798568.

[2] Karystianis G, Florez-Vargas O. Application of a rule-based approach to identify patient eligibility for clinical trials. In: proceedings of the 2018 National NLP Clinical Challenges (n2c2) Workshop Shared Tasks; January 28, 2018; San Francisco, CA.

[3] Xiong Y, Shi X, Chen S, Jiang D, Tang B, Wang X, Chen Q, Yan J. Cohort selection for clinical trials using hierarchical neural network. J Am Med Inform Assoc. 2019 Nov 1;26(11):1203-1208. doi: 10.1093/jamia/ocz099. PMID: 31305921; PMCID: PMC7647215.

[4] Author, J. A. (2020, November 12). An introduction to multi-label text classification [Blog post]. Analytics Vidhya. https://www.analyticsvidhya.com/blog/2017/08/introduction-to-multi-label-classification/

[5] Vashisht, A. (2019, October 27). *Luhn's Heuristic Method for text summarization*. OpenGenus IQ: Computing Expertise & Legacy. https://iq.opengenus.org/luhns-heuristic-method-for-text-summarization/